\documentclass[sigconf]{acmart}

\usepackage{booktabs}
\usepackage{multirow}
\usepackage{graphicx}
\usepackage{amsmath}
\usepackage{xcolor}
\usepackage{threeparttable}

\newcommand{\wiki}{\textsc{WikiText}}
\newcommand{\web}{\textsc{C4}}
\newcommand{\finmix}{\textsc{FinMix}}

\AtBeginDocument{%
  }

\setcopyright{none}
\renewcommand\footnotetextcopyrightpermission[1]{}
\begin{document}

\title[Damage Predicts Recovery]{Damage Predicts Recovery:\\When Calibration Data Matters in Compressing Financial LLMs}

% \author{Anonymous Authors}
% \affiliation{\institution{Anonymous Institution}\city{}\country{}}

% %%
% %% The "author" command and its associated commands are used to define
% %% the authors and their affiliations.
% %% Of note is the shared affiliation of the first two authors, and the
% %% "authornote" and "authornotemark" commands
% %% used to denote shared contribution to the research.
\author{Junyi Ye}
\orcid{0000-0002-6348-5207}
\affiliation{%
  \institution{School of Computing, Montclair State University}
  \city{Montclair}
  \state{New Jersey}
  \country{USA}
}
\email{yej@montclair.edu}

\author{Mengjia Yu}
\affiliation{%
  \institution{School of Computing, Montclair State University}
  \city{Montclair}
  \state{New Jersey}
  \country{USA}
}
\email{yum1@montclair.edu}

\author{Debapriya Hazra}
\affiliation{%
  \institution{School of Computing, Montclair State University}
  \city{Montclair}
  \state{New Jersey}
  \country{USA}
}
\email{hazrad@montclair.edu}

\author{Guiling Wang}
\affiliation{%
  \institution{Department of Computer Science, New Jersey Institute of Technology}
  \city{Newark}
  \state{New Jersey}
  \country{USA}
}
\email{gwang@njit.edu}
\begin{abstract}
Post-training quantization and pruning rely on a small calibration
corpus. Whether specialized domains such as finance require
domain-matched calibration data remains unsettled. We argue that the
answer depends on the task-level damage caused by compression rather
than on domain mismatch. If compression preserves the target
capability, changing the calibration corpus has little effect. If
compression causes large losses, task-formatted calibration can
recover part of the loss. We test this hypothesis across two model
families, six compression configurations, three token-matched
calibration corpora, and ten financial classification and numerical
question-answering tasks. The results support this hypothesis.
Quantization largely preserves task performance, and calibration
choice has little effect in this case. Pruning reduces numerical QA
accuracy by over 40 points. In these damaged settings, another
generic corpus does not help, while FinMix, a mixture of financial
task examples, recovers a large part of the loss. The link between
damage and recovery holds across model families and scales. These
findings support a practical rule. Measure task-specific compression
damage first, and construct specialized calibration data only when
the damage is large.
\end{abstract}


% %%
% %% The code below is generated by the tool at http://dl.acm.org/ccs.cfm.
% %% Please copy and paste the code instead of the example below.




% Remove ACM conference running headers/footers
\pagestyle{plain}

\maketitle

\section{Introduction}
Post-training compression makes large language models cheaper to
deploy, but many widely used quantization and pruning methods depend on
a small calibration corpus. In practice, this corpus is often generic
text such as WikiText or C4, even when the compressed model will be
used in a specialized domain~\cite{frantar2022gptq,lin2023awq,
xiao2022smoothquant,frantar2023sparsegpt,sun2023wanda}. Whether
domain- and task-matched calibration improves downstream performance
remains unsettled: prior results range from substantial gains to little
change or even degradation~\cite{williams2024impact,ji2025beware,
williams2025selfcalibration}. Consequently, practitioners lack a clear
rule for deciding when specialized calibration data is worth
constructing.

We argue that this question should not be answered from domain
mismatch alone. The relevant quantity is the task-specific
\emph{damage} caused by a particular model and compression method under
generic calibration. If compression already preserves the target
capability, changing the corpus leaves little room for improvement. If
it severely degrades that capability, task-formatted calibration may
recover part of the loss. Financial NLP provides a revealing test case
because its tasks range from label-based classification to numerical
question answering over reports and tables, allowing compression to
preserve one capability while severely degrading another.

We test this hypothesis in a controlled study of six compression
configurations, three calibration corpora, and ten financial tasks. The
main experiments use Llama-3.1-8B, while Qwen3-8B evaluates transfer to
a second model family and Llama-3.1-70B evaluates scale. For each
calibration-dependent method, we compare WikiText and C4 with
\finmix{}, a token-matched mixture of financial training examples in
the target-task format. Every compressed model is compared with its
BF16 reference on the same test examples using paired statistical
tests; free-form numerical answers are evaluated with a
human-validated LLM judge.

The results reveal a systematic damage--recovery relationship.
Quantization usually leaves task performance close to BF16, in which
case calibration choice has little effect. Pruning causes much larger
losses, particularly on numerical QA, and these losses are not reliably
reflected by generic-text perplexity. In the damaged settings,
\finmix{} recovers performance more consistently than replacing
WikiText with another generic corpus. Across Llama-3.1-8B and Qwen3-8B,
damage and recovery are strongly associated across 90 above-chance method–task observations (Spearman $\rho=0.74$). The relationship
also explains the smaller gains at 70B: the larger model suffers less
compression damage and therefore leaves less performance to recover.
Our results therefore suggest that calibration effort should be
matched to measured compression damage, not assigned solely because
the deployment domain is financial.

Our contributions are threefold:
\begin{itemize}
\item \textbf{Capability-selective damage.} We show that compression,
especially pruning, disproportionately damages financial numerical QA,
while classification is often preserved and generic perplexity can
miss the resulting loss.
\item \textbf{Damage-dependent recovery.}
We show that financial, task-formatted calibration is most effective
when generic calibration has caused substantial task-level damage.
A second generic-corpus control and repeated calibration draws confirm
that this recovery is not merely caused by changing the calibration
sample. Paired error analysis further shows that \finmix{} primarily
restores answers that the uncompressed model could already produce.
\item \textbf{A generalizable decision rule.} We show that task-level
damage is a strong predictor of the benefit of changing the
calibration corpus across the model families and scales tested,
supporting a practical damage audit before investing in specialized
calibration data.
\end{itemize}
The \finmix{} corpus, code, and all evaluation outputs will be made
publicly available upon acceptance.
\section{Related Work}

\textbf{Calibration data in post-training compression.}
Calibration-dependent methods such as GPTQ, AWQ, SmoothQuant,
SparseGPT, and Wanda typically use a small sample of generic or
pretraining-like text~\cite{frantar2022gptq,lin2023awq,xiao2022smoothquant,
frantar2023sparsegpt,sun2023wanda}. Their original studies focus on the
compression algorithms rather than comparing calibration corpora.
Williams and Aletras~\cite{williams2024impact} later showed that the
corpus can change downstream performance, particularly under pruning.
Subsequent work replaces random generic text with data closer to the
pretraining distribution~\cite{ji2025beware}, model-generated
text~\cite{williams2025selfcalibration}, or mixtures of generic and
domain- or task-specific examples~\cite{williams2026specialized,
wang2026alignment}. However, it remains unclear whether financial tasks require
domain-matched calibration or whether generic calibration is
sufficient.

\textbf{Evaluating compressed LLMs.}
Perplexity is commonly used to evaluate compression, but it does not
show whether individual capabilities remain intact. Jaiswal et
al.~\cite{jaiswal2024compressing} find that pruning can leave
perplexity nearly unchanged while substantially degrading
knowledge-intensive tasks. In their evaluation, quantization is more
robust. Liu et al.~\cite{liu2023emergent} further show that quantization
effects depend on both precision and capability: in-context learning,
chain-of-thought reasoning, and instruction following largely survive
at 4 bits but deteriorate sharply at 2 bits. These findings motivate
task-level evaluation rather than reliance on perplexity alone.

\textbf{Financial LLMs and benchmarks.}
BloombergGPT~\cite{wu2023bloomberggpt}, FinGPT~\cite{yang2023fingpt},
and PIXIU~\cite{xie2023pixiu} develop financial models through domain
pretraining or instruction tuning. PIXIU and FinBen~\cite{xie2024finben}
also consolidate benchmarks covering financial classification,
information extraction, and numerical reasoning. These studies focus
on building and evaluating financial LLMs. They do not examine how
post-training quantization and pruning affect these capabilities under
different calibration corpora.
\section{Experimental Setup}\label{sec:setup}

\subsection{Models and Compression Configurations}\label{sec:models}
The main experiments use Llama-3.1-8B. Qwen3-8B tests whether the
findings transfer to a second model family, and Llama-3.1-70B tests
whether they remain at a larger scale. We use the Instruct version of
each model and compare every compressed model with its corresponding
BF16 checkpoint.

We evaluate six compression configurations produced with
\texttt{llm-\\compressor}\footnote{\url{https://github.com/vllm-project/llm-compressor}}.
RTN, GPTQ~\cite{frantar2022gptq}, and AWQ~\cite{lin2023awq} use W4A16 quantization, while SmoothQuant~\cite{xiao2022smoothquant} uses
W8A8. The pruning configurations are SparseGPT~\cite{frantar2023sparsegpt} at 50\% unstructured
sparsity and Wanda~\cite{sun2023wanda} at 4:8 semi-structured sparsity. RTN does not use
calibration data and therefore serves as a calibration-free reference.
Each of the other five configurations is evaluated with the three
calibration corpora introduced next. GPTQ, SparseGPT, and Wanda use
corpus statistics to determine how weights are quantized or removed,
whereas AWQ and SmoothQuant use them to set scaling factors.

\subsection{Calibration Corpora}\label{sec:corpora}
Table~\ref{tab:calib} summarizes the three calibration corpora and
their roles. The comparison distinguishes a generic-corpus swap from a
switch to financial, task-formatted calibration data. \wiki{} is the
generic reference in the main damage analysis and follows the original
GPTQ setup. \web{} provides a second generic corpus and follows the
original AWQ setup. Comparing it with \wiki{} shows whether performance
changes after replacing one generic corpus with another. \finmix{} then
tests the combined benefit of financial content and task-style input
format.

\finmix{} is constructed from the training splits of the ten public
financial benchmarks listed in Table~\ref{tab:tasks}. We give each task the same token budget so that no single
benchmark dominates the mix. Each training example is converted into
the same input format used to evaluate its task. We remove any example
that duplicates an item in a test set.

All five calibration-based methods consume each corpus in full.
\wiki{} and \web{} each contain 128 sequences truncated to 2048
tokens, for a total of $262{,}144$ tokens. \finmix{} contains 2042 shorter
examples whose lengths vary by task and total approximately 260k
tokens. The corpora therefore match the amount of calibration text,
although they differ in the number and length of examples.

Prior work finds diminishing returns from adding calibration examples
at this scale and commonly uses 128 long sequences
\cite{frantar2023sparsegpt,sun2023wanda,williams2024impact}. The larger
number of \finmix{} examples should therefore not be read as a larger
calibration budget: it reflects the shorter length of task examples,
while the total number of tokens remains matched. The \finmix{}
comparison intentionally changes both financial content and input
format. We interpret its effect as their practical combination rather
than as a pure domain effect, since domain-specific examples normally
appear in the format of the target task.

\begin{table}[t]
\caption{Calibration corpora used in the experiments. All
calibration-based methods consume each corpus in full.}
\label{tab:calib}
\centering
\setlength{\tabcolsep}{4.5pt}
\small
\begin{tabular}{lllrr}
\toprule
Corpus & Content & Role & Seqs & Tokens \\
\midrule
\wiki{} & encyclopedia text & reference & 128 & 262k \\
\web{} & web text & generic control & 128 & 262k \\
\finmix{} & financial tasks & matched corpus & 2042 & 260k \\
\bottomrule
\end{tabular}
\end{table}

\subsection{Evaluation Tasks}\label{sec:tasks}
Table~\ref{tab:tasks} reports the size, label balance, numerical
dependence, and metric for each task. All ten come from public
financial benchmarks. We use the \texttt{flare} versions distributed
with PIXIU~\cite{xie2023pixiu} where available and evaluate on the
official test splits. The selected tasks span a broad range of
financial language understanding and reasoning capabilities, including
sentiment and stance classification, relation and causal understanding,
directional prediction, question answering, and multi-step numerical
reasoning.

\begin{table}[t]
\caption{Evaluation tasks. $n$ = canonical test-set size. Balance =
class distribution (\%); for FinRED (29 relation types) we report only
the largest class share, and the three free-form QA tasks have no
fixed label set (--). Num.\ = degree of numerical dependence.
Classification tasks report weighted F1, except FinCausal, which
reports macro-F1. QA tasks report
accuracy (judge-verified for the generative tasks, marked $\dagger$).}
\label{tab:tasks}
\centering
\setlength{\tabcolsep}{3pt}
\small
\begin{tabular}{llrlll}
\toprule
Task & Type & $n$ & Balance & Num. & Metric \\
\midrule
FPB~\cite{malo2014fpb} & 3-cls sentiment & 970 & 59/29/12 & none & wF1 \\
FOMC~\cite{shah2023fomc} & 3-cls stance & 496 & 49/26/25 & none & wF1 \\
Headlines~\cite{sinha2021headlines} & binary attribute & 1500 & 67/33 & low & wF1 \\
FinRED~\cite{sharma2022finred} & 29-cls relation & 1500 & max 13 & none & wF1 \\
FinCausal~\cite{mariko2020fincausal} & 2-cls causality & 1500 & 94/6 & low & mF1 \\
BigData22~\cite{soun2022bigdata22} & 2-cls movement & 1472 & 55/45 & low & wF1 \\
CFA~\cite{xie2023pixiu} & 3-way exam QA & 1032 & 35/34/31 & medium & Acc \\
FinQA~\cite{chen2021finqa} & numerical QA & 1147 & -- & high & Acc$^\dagger$ \\
ConvFinQA~\cite{chen2022convfinqa} & conv.\ numerical QA & 1490 & -- & high & Acc$^\dagger$ \\
TAT-QA~\cite{zhu2021tatqa} & table+text QA & 1500 & -- & high & Acc$^\dagger$ \\
\bottomrule
\end{tabular}
\end{table}

Six tasks use classification labels. FPB identifies the sentiment of a
financial sentence, and FOMC identifies the policy stance of a
central-bank statement. Headlines pairs a gold-market headline with
one of nine yes/no questions about attributes such as price direction,
time, events, or asset comparisons. FinRED identifies the relation
between two financial entities in context. FinCausal determines
whether a financial causal relation is expressed. BigData22 predicts
the next-day direction of a stock from tweets and recent price history.

Four tasks use question answering. CFA contains multiple-choice
questions about financial concepts and analysis. FinQA requires the
model to find values in financial reports and combine them through
multiple calculation steps. ConvFinQA adds conversational follow-up
questions whose answers can depend on earlier turns. TAT-QA asks for
calculated or extracted answers from evidence presented jointly as a
table and accompanying text.

\subsection{Evaluation and Scoring}\label{sec:protocol}
Every configuration of a task is scored on the same test examples and
the same prompts, and an automated check verifies this before
scoring. Decoding is greedy throughout, chat templates are fixed
within each model family, and Qwen3-8B runs with thinking mode
disabled.

Classification tasks are scored from label log-probabilities rather
than from generated text, so a degraded model cannot lose points for
formatting alone. Headlines compares the single tokens Yes and No.
FPB compares the three sentiment words, which tokenize to matched
lengths. The labels of FOMC, FinRED, FinCausal, and BigData22 tokenize
to unequal lengths, which direct scoring would reward, so their
choices are mapped to single-token option letters, the mapping is
rotated through every position, and the scores are averaged; the
rotation removes both the length advantage and any preference for a
particular letter or position. Each scorer verifies its tokenization
and option mapping at startup and stops rather than mis-score. CFA is
the exception: the model reasons freely and its final choice among
the three options is extracted, and a response with no extractable
choice counts as incorrect.

CFA and the three free-form QA tasks generate up to 2{,}048 tokens,
greedily. The QA tasks are scored by DeepSeek
(\texttt{deepseek-v4-pro}, temperature $0$, $4{,}096$-token limit),
which accepts equivalent numeric surface forms such as $0.146$ and
$14.6\%$, together with equivalent sign and unit conventions. The
judge prompt and settings are identical for every configuration, and
DeepSeek provides all reported scores on these tasks. Generations
that reach the token limit are marked incorrect without judging.

\subsection{LLM Judge Selection and Validation}\label{sec:judge-validation}

We validate our choice of LLM judge in three steps. First, an
independent second judge (Claude Sonnet 5) re-judges 3,000 responses
drawn from all judged configurations, blind to DeepSeek's verdicts.
The two judges agree on 91.6\% of them ($\kappa=0.76$).

Second, we compare both judges against human labels. An author labels
all 252 cases on which the judges disagree, plus 252 sampled cases on
which they agree, using written scoring rules and without seeing either
judge's verdict or the model configuration. After weighting the two
groups by their frequency, DeepSeek agrees with the human on 91.4\% of
verdicts (Cohen's $\kappa=0.83$), compared with 90.0\% for Claude.
Among cases where the judges disagree, the human agrees with DeepSeek
three times out of five. The two judges also show different error
patterns: DeepSeek tends to reject answers that the human accepts,
whereas Claude tends to accept answers that the human rejects. We
therefore use DeepSeek as the reporting judge because it is closer to
the human labels and is less likely to credit borderline answers. This
makes the reported recovery estimates conservative rather than
inflated.

Third, we re-score every configuration with Claude to test whether the
choice of judge affects our results. Absolute scores increase by four
to fifteen points, but all significant calibration contrasts remain
significant and retain the same direction. Our conclusions therefore
do not depend on the choice of judge.
% The annotations, scoring rules,
% and both sets of judge verdicts are released with the code.

\subsection{Statistical Analysis}\label{sec:statistics}

We test each calibration contrast separately for each task. For tasks
reported with accuracy, we apply an exact two-sided McNemar test to
paired correctness outcomes. For tasks reported with F1, we
resample paired predictions over test examples 2,000 times and
recompute F1 on each bootstrap sample. Stars indicate unadjusted task-level significance at the thresholds reported in the result tables.

WikiText test perplexity is computed over non-overlapping 2,048-token
windows. We also test the sampling stability of the pruning results by
repeating SparseGPT and Wanda with three independently drawn \wiki{}
calibration sets. Each run is compared with the same full \finmix{}
corpus.

\section{Degradation under Generic Calibration}\label{sec:risk}

% AUTO-GENERATED by scripts/analysis/generate_main_table_tex.py -- do not edit by hand
\begin{table*}[t]
\caption{Performance of \mbox{Llama-3.1-8B} under \wiki{}
calibration, with BF16 as the reference (RTN is calibration-free).
Higher task scores are better. Lower PPL is better. Avg.\ is the
unweighted mean within each task family. The parenthetical $\Delta$
gives the damage relative to BF16 (BF16 $-$ compressed), so positive
values indicate a loss. The classification mean excludes BigData22.
Stars denote significant differences from BF16 ($^{*}p{<}.05$).}
\label{tab:main}
\centering\small
\setlength{\tabcolsep}{0.6pt}
\begin{tabular}{llccccccccccccc}
\toprule
\multicolumn{3}{l}{Calibration: \wiki{}} & \multicolumn{7}{c}{\emph{Classification}} & \multicolumn{5}{c}{\emph{Numerical QA}} \\
\cmidrule(lr){4-10}\cmidrule(lr){11-15}
 & Method & PPL$\downarrow$ & FPB$\uparrow$ & FOMC$\uparrow$ & Headlines$\uparrow$ & FinRED$\uparrow$ & FinCausal$\uparrow$ & BigData22$\uparrow$ & \textbf{\emph{Avg.}}$\uparrow$ & CFA$\uparrow$ & FinQA$\uparrow$ & ConvFinQA$\uparrow$ & TAT-QA$\uparrow$ & \textbf{\emph{Avg.}}$\uparrow$ \\
\midrule
 & BF16 & 7.22 & 79.0 & 51.3 & 67.9 & 75.2 & 53.5 & 49.6 & \emph{65.4} & 68.1 & 43.9 & 64.7 & 74.1 & \emph{62.7} \\
\midrule
\multirow{4}{*}{Quantization} & RTN & 7.96 & 78.7 & 50.4 & 58.1$^{*}$ & 73.8 & 47.4$^{*}$ & 43.4$^{*}$ & \emph{61.7 ($\Delta$\,3.7)} & 62.4$^{*}$ & 39.6$^{*}$ & 59.1$^{*}$ & 69.5$^{*}$ & \emph{57.6 ($\Delta$\,5.1)} \\
 & GPTQ & 7.49 & 77.9 & 51.9 & 68.0 & 72.0$^{*}$ & 53.8 & 58.3$^{*}$ & \emph{64.7 ($\Delta$\,0.7)} & 62.8$^{*}$ & 36.9$^{*}$ & 58.6$^{*}$ & 70.5$^{*}$ & \emph{57.2 ($\Delta$\,5.5)} \\
 & AWQ & 7.62 & 76.4$^{*}$ & 49.2 & 68.2 & 74.0 & 52.7 & 49.3 & \emph{64.1 ($\Delta$\,1.3)} & 64.8$^{*}$ & 45.8 & 59.8$^{*}$ & 73.5 & \emph{61.0 ($\Delta$\,1.7)} \\
 & SmoothQuant & 7.33 & 79.0 & 53.1$^{*}$ & 67.7 & 73.7$^{*}$ & 54.3 & 50.5 & \emph{65.6 ($\Delta$\,$-0.2$)} & 66.8 & 44.8 & 63.1 & 73.0 & \emph{61.9 ($\Delta$\,0.8)} \\
\midrule
\multirow{2}{*}{Pruning} & SparseGPT & 9.57 & 77.7 & 45.4$^{*}$ & 23.7$^{*}$ & 67.1$^{*}$ & 44.9$^{*}$ & 39.9$^{*}$ & \emph{51.8 ($\Delta$\,13.6)} & 53.5$^{*}$ & 24.9$^{*}$ & 44.4$^{*}$ & 54.7$^{*}$ & \emph{44.4 ($\Delta$\,18.3)} \\
 & Wanda & 18.17 & 65.2$^{*}$ & 45.0$^{*}$ & 41.9$^{*}$ & 59.4$^{*}$ & 21.1$^{*}$ & 43.7$^{*}$ & \emph{46.5 ($\Delta$\,18.9)} & 32.9$^{*}$ & 6.7$^{*}$ & 16.6$^{*}$ & 31.3$^{*}$ & \emph{21.9 ($\Delta$\,40.8)} \\
\bottomrule
\end{tabular}
\end{table*}

RQ1 asks what compression damages in financial LLMs and whether the
field's standard checks would catch it. Table~\ref{tab:main} reports
every method under \wiki{} calibration against the BF16 reference on
all ten tasks. The results show two patterns. First, quantization
usually causes small losses, pruning causes much larger losses, and
numerical QA loses more than classification under every method.
Second, perplexity does not reliably show these differences across
tasks.

\subsection{Task-Selective Damage}\label{sec:selectivity}

For all six methods in Table~\ref{tab:main}, the average score drops
more on numerical QA than on classification. The size of the drop,
however, depends strongly on the compression method.
Among the quantization methods, SmoothQuant and AWQ leave both task
families largely intact, while RTN and GPTQ cause moderate losses that
fall more heavily on numerical QA. SparseGPT and Wanda produce much
larger losses, and numerical QA again suffers more than classification.
In short, numerical QA is more vulnerable under every method, and
pruning causes the largest overall decline. We exclude the
near-chance stock-movement task (BigData22) from the grouped classification mean,
because every model remains close to chance. Score changes mainly
reflect different guessing behavior rather than useful prediction
skill.

Overlong generation is another visible form of pruning damage. BF16 models typically reach the 2,048-token answer limit on fewer than 4\% of QA examples, while the most damaged pruning configuration reaches it on about 45\%. Inspection of the limit-hit outputs shows repeated or off-track reasoning without a final answer, rather than valid solutions cut short by the token budget. Reaching the limit therefore reflects a failure to terminate, not simply a need for a longer answer. These outputs remain in the evaluation set and are counted as incorrect.

GPTQ and RTN provide the clearest comparison among the quantization
methods. Both use the same four-bit weight format, but only GPTQ uses
calibration examples to guide weight reconstruction. GPTQ has better
average task performance than RTN, yet this advantage comes mainly
from classification. Its numerical-QA loss remains comparable to
RTN's. Thus, using calibration examples during reconstruction appears
to change which task family bears the loss, not only the total amount
of loss.

One plausible explanation is that reconstruction on generic text
better preserves the language understanding used in classification
than the processing required for multi-step numerical answers. This
comparison cannot establish that mechanism, because GPTQ and RTN
differ in more than their use of calibration data.
Section~\ref{sec:repair} therefore
holds the compression method fixed and changes only the calibration
corpus. If the explanation is correct, a corpus containing financial
examples and multi-step numerical questions should improve numerical
QA when generic calibration causes substantial losses, but should
matter little when compression causes little damage.

\subsection{Limits of Perplexity}\label{sec:ppl}

Perplexity measures how well a language model predicts the next token
in a text corpus. Lower values are better. Because
it requires no task labels, it is commonly used as a quick quality
check after compression. We compute it on held-out WikiText and ask
whether models with better perplexity also lose less performance on
the financial tasks in Table~\ref{tab:main}.

For classification, ranking the compressed models by WikiText
perplexity gives the same order as ranking them by their average score
loss. For numerical QA, it does not. GPTQ, for example, has better
perplexity than AWQ but worse performance throughout the numerical-QA
block. Better perplexity on generic text therefore does not imply
better numerical-QA performance.

A second limitation is that perplexity cannot show which downstream
capabilities compression has damaged. It assigns one overall score to
each model, while Table~\ref{tab:main} shows that the same compressed
model can retain classification performance and lose much of its
numerical capability. Wanda is the clearest example: it retains much
more classification performance while numerical QA largely collapses.
A single perplexity value cannot represent this difference.

\textbf{Answer to RQ1.} Under every method tested, numerical QA loses
more than classification, and pruning causes the largest absolute
losses. Perplexity ranks the methods correctly for classification but
not for numerical QA, and one perplexity value cannot show which tasks
have deteriorated.

\section{Recovery with Domain-Matched Calibration}\label{sec:repair}

\begin{figure*}[t]
\centering
\includegraphics[width=0.98\textwidth]{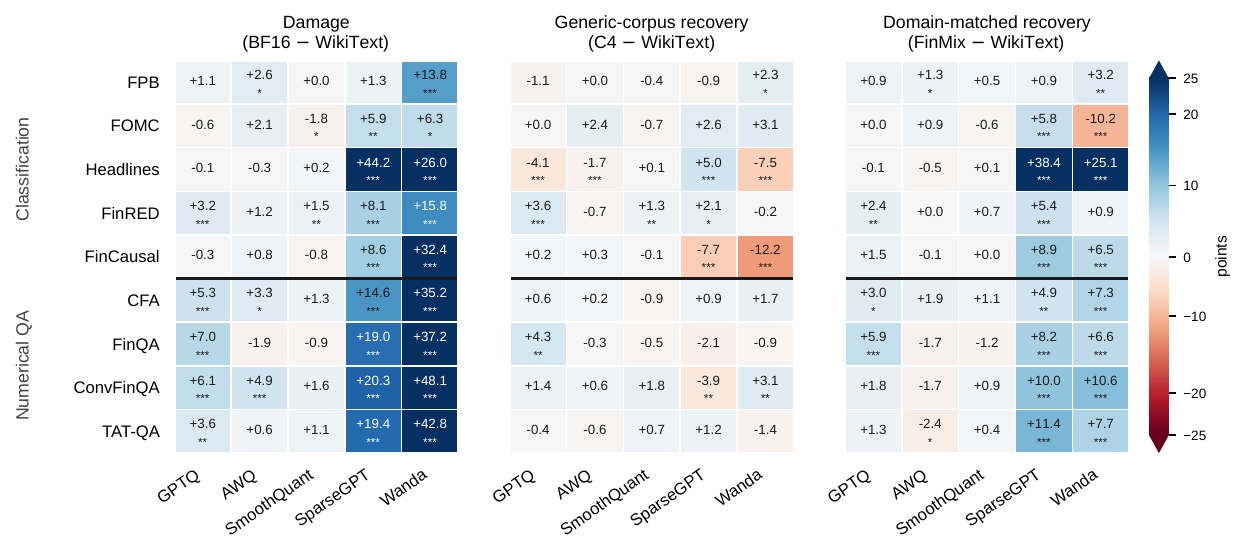}
\caption{Compression damage under \wiki{} calibration (left), and the
change after replacing \wiki{} with \web{} (middle) or \finmix{}
(right). Positive values indicate damage in the left panel and
recovery in the other two panels. Stars indicate paired-test
significance ($^{*}p{<}.05$, $^{**}p{<}.01$, $^{***}p{<}.001$).}
\label{fig:decomp}
\end{figure*}

RQ2 asks when domain-matched calibration data recovers performance
lost to compression. It also asks whether the improvement is specific
to \finmix{} or would arise after any change of calibration corpus. We
replace \wiki{} with \finmix{} at the same token budget, with no
retraining and no change to the compression method. Replacing \wiki{}
with \web{} provides a generic-corpus control. We use \emph{recovery}
for the measured gain in task score and \emph{repair} for the act of
correcting damage or an individual error.

\subsection{Recovery Beyond a Generic-Corpus Swap}

Figure~\ref{fig:decomp} compares the improvement from \finmix{} with
the change produced by switching between generic corpora. The left
panel shows the damage under \wiki{} calibration.
The middle and right panels then compare \web{} and \finmix{},
respectively, against the same \wiki{} baseline. The middle panel
reports \web{}$-$\wiki{}, and the right reports
\finmix{}$-$\wiki{}. Using the same baseline shows whether broad
improvements appear after either corpus replacement. Because
\finmix{} changes both financial content and task-style formatting,
the comparison measures their joint effect rather than a pure domain
effect.

SparseGPT and Wanda show the clearest difference between the two corpus changes. Replacing \wiki{} with \web{} produces a mixture of small gains and losses, with no consistent improvement. In contrast, replacing \wiki{} with \finmix{} improves most tasks under both pruning methods, showing a broad improvement that does not appear when moving from one generic corpus to another. It also reduces overlong generation by more than half in the configurations where this pruning failure is most severe.

The other methods behave differently. GPTQ suffers less initial damage
than pruning, and neither corpus replacement changes its average
performance by much. SmoothQuant and AWQ also change little, consistent
with their small losses under \wiki{} calibration. Thus, \finmix{} is
not uniformly beneficial. Its clearest advantage appears after severe
pruning damage.

Overall, \finmix{} produces more consistent recovery after pruning
than the generic-corpus control, but its effect depends on the
compression method. We next test whether the observed recovery depends
on the particular calibration examples sampled.

\subsection{Robustness Across WikiText Draws}\label{sec:seeds}

To check sampling stability, we repeat SparseGPT and Wanda with three
independently sampled \wiki{} calibration sets and compare each result
with the same full \finmix{} corpus. The check covers three
classification and three numerical-QA tasks.

% % Robustness of the GPTQ FinMix-WikiText contrast to sampling on BOTH sides.
% % Panel A: WikiText re-drawn (seeds 42/43/44), FinMix fixed.
% % Panel B: FinMix re-constructed (seeds 42/43/44), WikiText fixed.
% % All runs full-protocol; recomputed 2026-07-29 via calibration_significance helpers.
% \begin{table}[t]
% \caption{Robustness of the GPTQ \finmix$-$\wiki{} contrast (pp) to
% sampling on both sides. Panel A re-draws the \wiki{} sample under three
% seeds with the \finmix{} corpus fixed; Panel B re-constructs the
% \finmix{} task mix under three seeds with the \wiki{} draw fixed.
% FinQA, ConvFinQA, and CFA report accuracy with exact McNemar; FPB
% reports weighted F1 with paired bootstrap. Pooled tests combine
% discordant counts across the three seeds.}
% \label{tab:seeds}
% \centering
% \small
% \begin{tabular}{lcccc}
% \toprule
% Seed & FinQA & ConvFinQA & CFA & FPB \\
% \midrule
% \multicolumn{5}{l}{\emph{Panel A: \wiki{} draw re-sampled}} \\
% 42 & $+5.9^{***}$ & $+1.8$ & $+3.0^{*}$ & $+0.8$ \\
% 43 & $+10.0^{***}$ & $+3.3^{*}$ & $+1.5$ & $+1.3$ \\
% 44 & $+4.6^{**}$ & $+2.3$ & $+1.3$ & $+0.2$ \\
% pooled $p$ & $2.5{\times}10^{-14}$ & $1.0{\times}10^{-3}$ & -- & -- \\
% \midrule
% \multicolumn{5}{l}{\emph{Panel B: \finmix{} construction re-sampled}} \\
% 42 & $+5.9^{***}$ & $+1.8$ & $+3.0^{*}$ & $+0.8$ \\
% 43 & $+3.8^{*}$ & $+0.5$ & $+2.7$ & $+0.3$ \\
% 44 & $+6.9^{***}$ & $+1.1$ & $+1.5$ & $+1.4$ \\
% pooled $p$ & $8.1{\times}10^{-10}$ & $0.12$ & -- & -- \\
% \bottomrule
% \end{tabular}
% \end{table}
\begin{table}[t]
\caption{Pruning recovery across three \wiki{} draws. Entries give
significantly positive comparisons over all task--seed comparisons
($p{<}.05$). All 36 recovery values are positive.}
\label{tab:seeds}
\centering
\small
\setlength{\tabcolsep}{10pt}
\begin{tabular}{lrrr}
\toprule
Method & \emph{Classification} & \emph{Numerical QA} & Overall \\
\midrule
SparseGPT & 8/9 & 9/9 & 17/18 \\
Wanda     & 9/9 & 9/9 & 18/18 \\
\bottomrule
\end{tabular}
\end{table}

Table~\ref{tab:seeds} shows that recovery is positive in all 36
method--task--seed comparisons and significant in 35. Every
numerical-QA comparison is significantly positive for both pruning
methods. The only nonsignificant result is one SparseGPT
classification run. The pruning recovery is therefore stable across
different \wiki{} calibration samples, with the strongest consistency
on numerical QA.

\subsection{Which Errors Are Repaired}\label{sec:repair-origin}

To determine whether \finmix{} restores performance lost during
compression, we examine every error made by GPTQ and SparseGPT under
\wiki{} calibration on FinQA and ConvFinQA. We divide these errors into
two groups. In the first, BF16 answers the sample correctly but the
compressed model does not. These errors were introduced by compression.
In the second, both models answer incorrectly. These errors were
already present in BF16.

\begin{table}[t]
\caption{\finmix{} repair rates among errors under \wiki{}
calibration, grouped by whether BF16 answers the same sample correctly.
The higher rate for BF16-correct samples indicates recovery of
performance lost during compression.}
\label{tab:origin}
\centering\small
\setlength{\tabcolsep}{7pt}
\begin{tabular}{lrrr}
\toprule
& \multicolumn{2}{c}{\shortstack{\finmix{} repair rate (\%)}} & \\
\cmidrule(lr){2-3}
Method & \shortstack{BF16 correct} & \shortstack{BF16 wrong} & \shortstack{OR {}[95\% CI]} \\
\midrule
GPTQ & 62.1 & 15.2 & 9.08 [6.95, 11.86] \\
SparseGPT & 48.0 & 16.1 & 4.62 [3.69, 5.80] \\
\midrule
Overall & 53.2 & 15.6 & 6.07 [5.11, 7.22] \\
\bottomrule
\end{tabular}
\begin{tablenotes}
\footnotesize
\item OR denotes the odds ratio. The GPTQ and SparseGPT estimates are
stratified by task. The Overall row combines both methods and is
stratified by method and task.
\end{tablenotes}
\end{table}

Table~\ref{tab:origin} shows the same pattern for both methods.
\finmix{} repairs about half of the errors introduced by compression,
but only about one in six of the errors already present in BF16. Thus,
an error is far more likely to be repaired when BF16 originally
answered the sample correctly. Combining both methods gives an odds
ratio of 6.07, with a 95\% confidence interval of $[5.11, 7.22]$.
This means that the odds of repair are about six times higher for
errors introduced by compression than for errors already made by BF16.
\finmix{} can occasionally solve a sample that BF16 misses, but it is
much more effective at recovering performance lost during compression.

To examine whether recoverability differs by error type, we manually
assign each compression-introduced error to one of the six categories
in Table~\ref{tab:errortype}. We then compare repair rates within GPTQ
and SparseGPT.

% AUTO-GENERATED by scripts/analysis/generate_error_type_repair.py -- do not edit by hand
\begin{table}[t]
\caption{Counts and \finmix{} repair rates for manually categorized,
compression-introduced errors across the two numerical tasks. The
Errors columns give category counts, and the Repair columns give the
percentage corrected by \finmix{}. Overall combines GPTQ and
SparseGPT.}
\label{tab:errortype}
\centering\small
\setlength{\tabcolsep}{1.5pt}
\begin{tabular}{lrrrrrr}
\toprule
& \multicolumn{2}{c}{GPTQ} & \multicolumn{2}{c}{SparseGPT}
& \multicolumn{2}{c}{Overall} \\
\cmidrule(lr){2-3}\cmidrule(lr){4-5}\cmidrule(lr){6-7}
Error type & Errors & Repair (\%) & Errors & Repair (\%) & Errors & Repair (\%)\\
\midrule
reasoning & 184 & 63 & 249 & 51 & 433 & 56 \\
unit/scale & 106 & 66 & 106 & 48 & 212 & 57 \\
arithmetic & 69 & 52 & 256 & 43 & 325 & 45 \\
extraction & 32 & 62 & 89 & 48 & 121 & 52 \\
format & 23 & 61 & 9 & 89 & 32 & 69 \\
other & 5 & 80 & 3 & 33 & 8 & 62 \\
\midrule
Repair-rate test $p$ & -- & 0.50 & -- & 0.086 & -- & -- \\
\bottomrule
\end{tabular}
\end{table}

\begin{figure*}[t]
\centering
\includegraphics[width=0.95\textwidth]{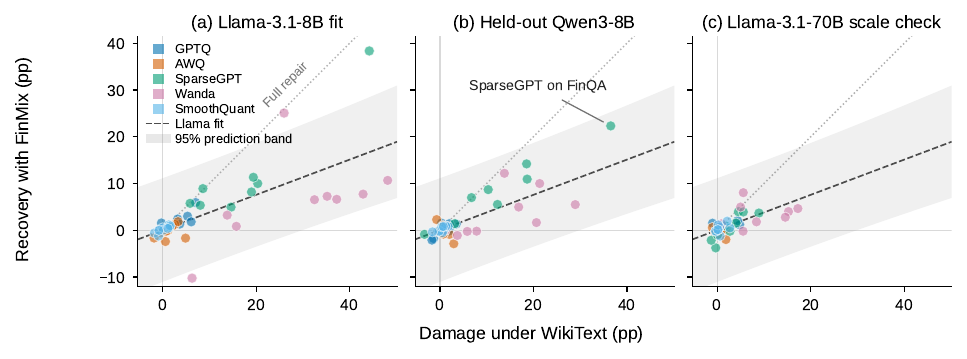}
\caption{Compression damage and recovery across model families and
scales. Panel (a) fits Llama-3.1-8B, panel (b) tests held-out
Qwen3-8B, and panel (c) checks Llama-3.1-70B. The black dashed line
and shaded area show the fit and its 95\% prediction band. The gray
dotted line marks full repair.}
\label{fig:law}
\end{figure*}

The error counts first show what compression breaks most often.
Reasoning and arithmetic together account for about two thirds of the
errors in this analysis. GPTQ produces relatively more reasoning and
unit-or-scale errors, while SparseGPT produces relatively more
arithmetic and extraction errors. Format and other errors are rare.

The repair columns answer a different question. They show the share of
each error type corrected by \finmix{}. Across the four common error
types, GPTQ has higher repair rates than SparseGPT, matching the
method-level difference in Table~\ref{tab:origin}. Within either method,
however, the differences across error types are not statistically
reliable. The high rates for the rare format and other categories are
based on very small counts and should not be interpreted as a stable
pattern. Thus, error type describes what compression tends to break,
but does not reliably predict which errors \finmix{} will repair.

\textbf{Answer to RQ2.} \finmix{} provides more consistent recovery
than a second generic corpus after severe pruning damage. Its effect is
modest for methods with less initial damage. Errors introduced by
compression are far more likely to be repaired than errors already
present in BF16. Among those compression-introduced errors, the
manually assigned error type does not reliably predict repair.

\section{Generalization Across Model Families and Scales}\label{sec:predict}

RQ3 asks whether the link between compression damage and recovery is
limited to Llama-3.1-8B. We test two kinds of generalization. Qwen3-8B
changes the model family while keeping a similar scale.
Llama-3.1-70B changes the scale while keeping the Llama family.

For each method and task, we define \emph{Damage} (D) as
BF16$-$\textsc{Wiki-\\Text} and \emph{Recovery} (R) as \finmix{}$-$\wiki{}.
Positive D means that compression lowers the score under \wiki{}
calibration. Positive R means that switching to \finmix{} raises it.
Figure~\ref{fig:law} plots D against R. Each panel contains 45 points
from five methods and nine above-chance tasks. Values are in percentage
points, and colors identify compression methods.

The black dashed line is estimated only from panel (a). Its shaded
band shows where new results would usually fall if they follow the
same relationship. The gray dotted line is full repair, where the gain
from \finmix{} equals all performance lost under \wiki{}. Points below
this line recover only part of the loss. The near-chance BigData22
task is excluded for the reason discussed in Section~\ref{sec:selectivity}.

\subsection{Generalization to Qwen3-8B}\label{sec:qwen}

Figure~\ref{fig:law}(a) uses only Llama-3.1-8B. Its fitted slope is
0.38, so each
additional point of damage is associated with about 0.38 additional
points of recovery. Methods with little damage remain near the origin,
while pruning produces both larger losses and larger gains from
\finmix{}. The most severely damaged Wanda points remain well below
the full-repair line. FinMix therefore recovers only part of the
performance lost under extreme pruning.

Figure~\ref{fig:law}(b) changes the model family but keeps the 8B
scale. Its Qwen3-8B points are not used to estimate the line or
prediction band.
Quantized Qwen3-8B remains near the origin, with little damage and
little response to corpus choice. Pruning moves the points outward.
The annotated SparseGPT result illustrates the pattern. FinQA suffers
substantial damage under \wiki{} calibration and recovers strongly
under \finmix{}. All 45 Qwen3-8B points fall within the Llama-3.1-8B
prediction band.

Across panels (a) and (b), damage and recovery have a Spearman
correlation of $\rho=0.74$ over 90 points. Including the near-chance
stock-movement task gives $\rho=0.68$ over all 100 points. The positive
relationship therefore does not depend on that exclusion, and the
pattern fitted on Llama-3.1-8B generalizes to a second model family.

\subsection{Effect of Model Scale}\label{sec:scale}

\begin{table}[t]
\caption{Average BF16 performance and compression damage/recovery
(D~/~R, pp) by method and model. Higher BF16 scores are better.}
\label{tab:summary}
\centering
\small
\setlength{\tabcolsep}{7pt}
\begin{tabular}{lrrr}
\toprule
Method & \shortstack{Llama-3.1-8B} & \shortstack{Qwen3-8B}
& \shortstack{Llama-3.1-70B} \\
\midrule
\multicolumn{4}{l}{\emph{Classification}} \\
BF16 score  & 65.4          & 66.8         & 70.7 \\
GPTQ        & $+0.7/+0.9$   & $-0.3/-0.4$  & $+0.8/+0.6$ \\
AWQ         & $+1.3/+0.3$   & $-0.1/+0.3$  & $+0.2/-0.5$ \\
SmoothQuant & $-0.2/+0.1$   & $+0.1/0.0$   & $+0.6/+0.2$ \\
SparseGPT   & $+13.6/+11.9$ & $+3.5/+3.3$  & $+1.0/-0.4$ \\
Wanda       & $+18.9/+5.1$  & $+6.5/+2.1$  & $+5.5/+3.8$ \\
\midrule
\multicolumn{4}{l}{\emph{Numerical QA}} \\
BF16 score  & 62.7          & 74.7          & 77.3 \\
GPTQ        & $+5.5/+3.0$   & $+2.3/+0.7$   & $+1.9/+1.0$ \\
AWQ         & $+1.7/-1.0$   & $+1.6/-0.9$   & $+0.8/-0.3$ \\
SmoothQuant & $+0.8/+0.3$   & $+0.7/+0.5$   & $+0.4/+0.1$ \\
SparseGPT   & $+18.3/+8.6$  & $+21.5/+13.3$ & $+5.3/+2.2$ \\
Wanda       & $+40.8/+8.1$  & $+22.0/+5.6$  & $+11.4/+2.3$ \\
\bottomrule
\end{tabular}
\end{table}

Table~\ref{tab:summary} summarizes the same D and R values by task
group. Each entry is an average within the classification or
numerical-QA block. BigData22 is again excluded from classification.

The BF16 baselines show different cross-model patterns for the two
task groups. Classification performance varies only modestly.
Llama-3.1-70B performs best, while Qwen3-8B is only slightly above
Llama-3.1-8B. Numerical QA separates the models more clearly.
Qwen3-8B is much stronger than Llama-3.1-8B and approaches the 70B
model. Numerical performance therefore depends on model family as
well as scale.

The classification block is stable under quantization. GPTQ, AWQ,
and SmoothQuant remain close to BF16 on all three models, leaving
little for \finmix{} to recover. The larger differences come from
pruning. SparseGPT and Wanda cause substantial classification damage
on Llama-3.1-8B, but less on Qwen3-8B and Llama-3.1-70B. Recovery also
depends on the pruning method. \finmix{} removes most of the SparseGPT
loss at 8B, whereas the recovery from Wanda is more limited.

Numerical QA shows the clearest scale effect. Quantization remains
fairly stable, apart from the larger GPTQ loss on Llama-3.1-8B. Under
pruning, Llama-3.1-70B loses much less than either 8B model. The same
reduction appears for both SparseGPT and Wanda, suggesting that the
larger model is more tolerant of pruning. One possible explanation is
greater parameter redundancy at 70B, although our experiments do not
test that mechanism directly. Qwen3-8B shows that a strong BF16 score
does not provide the same protection. Its numerical performance is
close to Llama-3.1-70B before compression, yet it suffers much larger
pruning losses, including the largest SparseGPT loss. \finmix{}
recovers part of these losses, but does not fully restore the pruned
models.

Panel (c) in Figure~\ref{fig:law} shows the same comparison task by task. All 45 points fall
within the Llama-3.1-8B prediction band. Most are closer to the origin
than the 8B pruning points because compression causes less damage at
70B. With less damage, changing the calibration corpus also produces
a smaller gain. 
% Complete task-level results are released with the
% code.

\textbf{Answer to RQ3.} Damage under generic calibration remains a
useful indicator of the benefit of \finmix{} across model families and
scales. The Llama-3.1-8B fit generalizes to held-out Qwen3-8B, and the
70B results follow the same direction with smaller damage and smaller
recovery. The relationship generalizes, but its magnitude depends on
the model and scale.

\section{Practical Guidance}\label{sec:guidance}

Our findings suggest a three-step procedure for deploying compressed
models in financial NLP.

\begin{enumerate}
\item \textbf{Compare the compressed model directly with BF16.}
Run both models on the same representative validation examples and
report classification and numerical QA separately. If the application
requires calculations or reasoning over tables, include such examples
in this check. A combined score or generic perplexity can hide the
numerical damage.

\item \textbf{Choose the response based on the observed score loss.}
If the loss is already small, changing the calibration corpus is
unlikely to help much. If pruning causes a large loss, recalibrate with
task-formatted financial data. Replacing one generic corpus with
another did not provide the same reliable recovery. If the remaining
loss is still unacceptable, reduce the pruning level or choose a less
damaging compression method.

\item \textbf{Repeat the comparison after changing the model or
scale.} Qwen3-8B approaches the 70B model in uncompressed numerical
performance but suffers much larger pruning losses. A strong BF16
score therefore cannot replace a compression test on the new model.
Use the newly measured damage to judge whether changing the calibration
data is likely to help.
\end{enumerate}

\section{Conclusion}
This work reframes calibration-data selection as a response to
compression damage rather than a fixed preprocessing choice. The
question is not whether domain-matched calibration is always better.
It is whether compression has degraded capabilities that better
calibration data can recover. When little has been lost, changing the
corpus has little value. When a capability has been severely damaged,
task-formatted calibration can recover part of the loss. This
distinction matters for financial LLMs in particular, because
aggregate scores and generic language-model metrics can hide losses
in numerical reasoning. Calibration should therefore follow a
task-level assessment of what compression has changed.

Our study has limitations. \finmix{} changes both financial content
and task format, so the two effects are not separated. The scale
analysis uses a single 70B model, and the mechanism behind its
smaller losses is not tested directly. Whether the damage-first rule
extends to other specialized domains remains an open question. Within
the settings tested, however, measured task-level damage provides a
clearer basis than domain membership for deciding when specialized
calibration data is worth the effort.

\bibliographystyle{ACM-Reference-Format}
\bibliography{references}

\end{document}